# Interactive Tools and Tasks for the Hebrew Bible: From Language Learning to Textual Criticism[1]

**Nicolai Winther-Nielsen**

Fjellhaug International University College Denmark

Correspondence address author: nwn@dbi.edu

**Abstract**
This contribution to a special issue on "Computer-aided processing of intertextuality" in ancient texts will illustrate how using digital tools to interact with the Hebrew Bible offers new promising perspectives for visualizing the texts and for performing tasks in education and research. This contribution explores how the corpus of the Hebrew Bible created and maintained by the Eep Talstra Centre for Bible and Computer can support new methods for modern knowledge workers within the field of digital humanities and theology be applied to ancient texts, and how this can be envisioned as a new field of digital intertextuality. The article first describes how the corpus was used to develop the Bible Online Learner as a persuasive technology to enhance language learning with, in, and around a database that acts as the engine driving interactive tasks for learners. Intertextuality in this case is a matter of active exploration and ongoing practice. Furthermore, interactive corpus-technology has an important bearing on the task of textual criticism as a specialized area of research that depends increasingly on the availability of digital resources. Commercial solutions developed by software companies like Logos and Accordance offer a market-based intertextuality defined by the production of advanced digital resources for scholars and students as useful alternatives to often inaccessible and expensive printed versions. It is reasonable to expect that in the future interactive corpus technology will allow scholars to do innovative academic tasks in textual criticism and interpretation. We have already seen the emergence of promising tools for text categorization, analysis of translation shifts, and interpretation. Broadly speaking, interactive tools and tasks within the three areas of language learning, textual criticism, and Biblical studies illustrate a new kind of intertextuality emerging within digital humanities.

## INTRODUCTION

In her 9th annual list of *Top 100 Tools for Learning 2015*, Jane Hart concludes from the response of more than two thousand top educators that again the list is "dominated by free online (social) tools" [http1]. In her analysis of these surveys, [Hart, 2014: 25] furthermore distils five key features profiling how high performing knowledge workers like to learn in flow, in an ongoing manner, on demand, socially, and autonomously. This hardly matches the traditional profile of scholars specializing in the study of ancient texts. Yet modern universities who are educating students for the new global society no doubt are also raising the professional bar for education and research in their Oriental or Biblical Studies departments, not the least because these departments typically educate students for specialized low-income jobs. These students even often have to assume jobs outside their narrow field of specialization. In this process of inevitable change for modern social learning, the creation of databases for the study of ancient texts like the Hebrew Bible can play a key pedagogical role. I will make a case for using corpora to train knowledge workers to use new tools for new tasks. In particular, I will discuss this kind of digital interaction for a corpus of the Hebrew Bible constructed and maintained by the Eep Talstra Centre for Bible and Computer (ETCBC) at the VU University in Amsterdam. Starting in 1977, for some forty years researchers in funded projects and dissertations have developed a major scholarly resource for morphological and syntactical analysis of the Hebrew Bible. The ETCBC database is now openly accessible: one may download it or use it online through the SHEBANQ project, short for «System for HEBrew Text: ANnotations for Queries and Markup» [http2], especially designed to allow users to annotate the text with queries.

---

[1] I would like to express my sincere thanks to Randall Tan for editing my English, to my students, Christian Højgaard Jensen and Conrad Thorup Elmelund, and to respondent Carsten Vang for discussing the paper at the Old Testament research session at Helsjön.

As the corpus itself is well-documented online through annotations and reference resources [http2], this contribution will focus on how the database can be applied in linguistic projects, illustrating its usefulness for computer-aided processing of intertextuality. The paper endeavours to define a new sense of intertextuality, not in terms of the traditional use of this term in literary criticism and rhetorical studies, but in terms of interaction with digital repositories of ancient corpora. This contribution presents different angles on the role of the intertextuality of the Hebrew Bible in three sections. First, I describe how the ETCBC already functions as a tool for task-based training of crucial skills in language learning. The second section shifts to the challenge of textual criticism in a discussion of digital tools based on the ETCBC that are already commercially available for interactive study and education. The third section endeavours to look at the challenge of computer-driven interactivity for the future by proposing a new framework for what I will call "textual corpus criticism", envisioning new tools and tasks for corpus-based studies of the Hebrew Bible.

**I. BIBLE ONLINE LEARNER: TASKS IN INTERACTIVE LEARNING**

Digital media are rapidly expanding the pace and space of social interaction and self-directed learning, fundamentally changing practice and outcomes for Second Language Acquisition. As an example, the dissertation of [Marissa, 2013] shows how intertextuality can enhance literacy practice for Indonesian English language learners using Twitter. Unfortunately, this kind of intertextuality will not work for the one-directional information processing of ancient languages like Biblical Hebrew and texts like the Hebrew Bible, as social interaction is gone for good. Digital media can help students converse in spoken Ivrith (Modern Hebrew), thus cultivating vocabulary acquisition, but this is a remix of modern language use and the study of ancient texts based on an artificial intertextuality, missing the full authentic reality of the world of the ancient documents.

As a substitute for the missing communication partner who disappeared thousands of years ago, we can still learn from their texts, and [Flowerdew, 2012] shows the great potential of corpora for language learning. At the outset, this requires that the texts in the corpus are both teacher-effective and user-friendly. I, therefore, first focus on two areas of the 'Learning Object Rating Instrument' proposed by [Nesbit and Belfer, 2004: 148]: the corpus must be motivational, providing the "[a]bility to motivate and stimulate the interest or curiosity of an identified population of learners". It must also afford interaction usability, supporting "[e]ase of navigation, predictability of the user interface and the quality of User Interface help features". In the following, I will describe various persuasive models designed to foster ability and motivation. Then I will explain the design of a learner-friendly interface for the corpus. Lastly, I will provide global evidence for the successful operation of this system.

## 1.1 Models for Corpus-Driven Persuasive Learning

When I joined the research-team of the ETCBC some 25 years ago, I soon discovered how effective it was to integrate a strong corpus-technology in the analysis of the linguistics, discourse structure, and interpretation of a book of the Hebrew Bible like Joshua [Winther-Nielsen, 1995]. Starting in 2003, I taught Biblical Hebrew and in collaboration with Dr Ulrik Sandborg-Petersen implemented an interactive quiz tool for practicing all essential paradigms of Biblical Hebrew.[2] Observing how students acquired new skills for grammatical analysis of the tricky Hebrew paradigms, I began wondering how we could turn the entire corpus of the Hebrew Bible into an interactive task for the learner and thus build on the impressive learning results already obtained [Winther-Nielsen, 2011]. The latter was made possible by a grant to explore the effect of persuasion for corpus-based learning in the EU project EuroPLOT 2010-2013.[3]

From the beginning, our design for learning with, in, and around a corpus was based on the ideas developed for Persuasive Technology by [Fogg, 2003]. His theory provided an approach to a design with a focus on a tool for training, on a medium for simulation, and on the presence of a social actor. In my first model, I sorted his seven functions of a tool according to whether they support ability or motivation and as to how effective the functions are on a cline.[4] Fogg's first three functions support ability. Most learners first need to be enabled to perform simple tasks by (1) the reduction of the complexity of a learning task. They may then proceed to (2) tunnelling, which takes the user step-by-step through acquiring some skills chosen by a teacher. Then (3) tailoring can adapt to an individual learner's needs based on a calculation of the goals and achievements of other learners. The rest of the functions support motivation.[5] The crudest way is motivation through (7) conditioning, like when rewards are provided in a self-corrective exam. A more persuasive motivation is through (6) surveillance, where teachers supervise learners and monitor their learning outcomes continuously. However, true internalized motivation can develop through (5) self-monitoring when learners can direct their own learning projects and define their own goals while observing what they do and reflect on how to improve their performance. The highest degree of motivation is reached when the technology simply offers suggestions at the right time. Thus the ultimate ideal learning state is reached in function (4) suggestion, when learners float around in a fine-tuned learning environment that matches the right kind of enablement and motivation at each appropriate moment, providing the optimal scaffolding for all tasks and outcomes.[6]

In all learning, it is crucial that students get instant response with corrective feedback. I therefore worked out a second model of instant feedback for corpus-driven learning from [Laurillard, 2012: 60] who has set up a framework for learning from a practice environment. In this kind of system, the corpus and additional teaching can model the learning, and the corpus also checks for errors made and instantly suggests corrections needed for further training.[7] Laurillard's model can integrate the social world into a design for learning and it explains how learners engage with other learners in a plotted practice environment.

---

[2] Ulrik Sandborg-Petersen has now developed this project into a commercial quiz tool for not only Hebrew, but also New Testament Greek and modern Spanish, as well as soon also Latin, see [http3].

[3] See the results published by [Behringer and Sinclair, 2013]. PLOT is an acronym for Persuasive Learning Objects and Technologies; EuroPLOT was funded by the Education, Audiovisual and Culture Executive Agency (EACEA) of the European Commission through the Livelong Learning Program with grant #511633.

[4] The seminal ideas were submitted by [Winther-Nielsen, MS] in 2012 for a special issue which unfortunately in 2016 has not yet been published. This first model is also summarized in [Winther-Nielsen, 2014: 84-85].

[5] Note that the numbering is not consecutive because the functions are listed according to Fogg's numbering of the seven functions rather than my sorted order.

[6] The suggestion is "the ultimate goal of any persuasive technology, at the peak of its persuasive force, the *khairos*" [Winther-Nielsen, 2014: 85].

[7] See [Winther-Nielsen, 2013a: 23-25], and [Gottschalk and Winther-Nielsen, 2013: 112-113].

The third and latest unified model [Winther-Nielsen, 2014: 85-87] did not only introduce a model to explain the force of a persuasive learning technology,[8] but – more importantly – it also proposed a design for flow in a RAMP-model, which handles the triggering mechanisms needed in a persuasive system. The goal in this proposal is to capture research on intrinsic motivation by Csikszentmihalyi, Deci, Ryan, Pink, and other experts in educational psychology. The model combines Relatedness, Autonomy, Mastery, and Purpose, hence the acronym RAMP. It is designed to bring learners into a state of mind where they are completely absorbed in their learning activities, where they love their learning challenges and forget about time and place. For the modern knowledge worker this means to learn through workflow, in an ongoing manner, where the learning is socially supportive, offering them a high degree of autonomy.[9]

Accordingly, the RAMP model starts with a crucial challenge for initial engagement which to some extent must come across as an invitation to each individual.[10] After commitment, [Winther-Nielsen, 2014: 87] envisions that learning technology will generate the optimal persuasive flow by "gradually strengthening the autonomy and increasing the mastery towards full self-directed control and perfection." Through appropriate instructional content and the right kind of ability support, the technology assists the learner in proceeding in the direction personally most desired. The seven functions for tools in Persuasive Technology are triggered at the right stage and place, creating a flow, through a learning experience that takes learners into a relationship defined by some social context.[11] In this sense, persuasive learning fuels enablement and motivation leading to mastery and autonomy, and offering social and professional relationships; together this "gives us the triggers that can ramp-up persuasive learning" [ibid: 87].

This latest version of the model of persuasive corpus-driven learning also includes a model for different kinds of contexts, layered as ever expanding ellipses around the RAMP. The innermost ellipsis is the context of the database as a macro learning object that can be programmed to function as a corpus-driven motor for learning. Around this object are the learners with all their own personal learning projects, while developers and learning designers are trying to create spaces for self-directed learning. The learner's practice environment is framed by a third and broader context, the educational environment, and this is where the purposes for engaging in learning are influenced by teachers and institutions. The fourth and widest kind of context includes all the informal social learning going on in society, and the group performance described by [Hart, 2014] for the knowledge worker. These factors define a broad kind of intertextuality that plays a role in the study of ancient texts, but the factors are fluid and not easy to pinpoint, because they are formed by the many roles of the Bible in church and society. These contexts form the four expanding spaces of learning objects, self-direction of learners, institutional facilitation, and peer-collaboration in a social world that must be taken into account in the learning environment.

---

[8] The distinction between the intended goal of the designer and the actual outcome is explained by [Winther-Nielsen, 2014: 83-84] through Speech Act Theory: "pedagogical content (pC) can fulfill an intended persuasive force (pF) under specific technological conditions. … the successful persuasive outcome (pO) … is foremost a matter of different kinds of contextual responses achieved by the persuasive event."

[9] See [Hart, 2014: 25]: (1) Knowledge workers learn in the flow of work, not in classes or online courses of long duration. (2) They learn continuously, by informally reading, overhearing, and observing (this accounts for some 80-90% of total learning). (3) They learn on demand, encountering information immediately. (4) They learn socially, working among peers (5) They learn autonomously, by self-directed, self-organised and self-managed choice and control."

[10] In the opinion of [Winther-Nielsen, 2014: 87], the efficiency of a technology depends on "its ability to adapt to all learners and their cultures, even to those who are forced to use the technology or have to force themselves to do so".

[11] According to [Winther-Nielsen, 2014: 87] the ultimate goal is to "function in some social context of a class, group or online community, and ultimately help the learner to achieve a social position at work and in society.

It may well be possible to refine these three persuasive models for interactivity around a database and come up with new explanations as to how force, flow, and context trigger persuasive learning. On the social end, our models have to be open and flexible in order to be able to accommodate all modern and global contexts that we now know and that may evolve in the future.

**1.2 The Learner-Friendly Corpus Interface in Bible Online Learner**

The models for interactive learning described above have already been implemented. We have a well-developed solution for learning from ancient texts that illustrates that our models are up and running.

The architecture of a tool for corpus-driven learning started with a project described by [Tøndering, 2009] as the PC program Ezer Emdros Exercise Tool (3ET) for quizzing using the ETCBC corpus. While 3ET applied the idea of reusable and repurposable learning objects well, it did not support motivation, and the program did not catch on. Fortunately, in the EU project we had the opportunity to redesign 3ET into a persuasive PC program, PLOTLearner. In 2013, we started to build further on the EU project, even though EU funding had ceased. We developed an online version for the PC program because we realized that it is difficult to develop efficiently for different platforms, and apps are not useful for large amounts of learning content such as the Hebrew Bible.[12] The new online tool is called Bible Online Learner, or Bible OL. It uses the ETCBC database and is interlinked with SHEBANQ, a second major tool.[13]

The role of this database within the studies of the humanities has already been described in [Sandborg-Petersen, 2011]. In order to appreciate how an ancient text is handled in a digital tool, it is important to understand the distinction between a database and a corpus. Developers of tools will have to construct and use a linguistically annotated text-database as a system for storing and retrieving data and for querying the annotated data as described in [Sandborg-Petersen, 2011: 263], but this database only serves as a corpus when "its primary function is to be a research instrument" . The key to the new interactive way of learning from an ancient text is the ability to design for learning with, in, and around this corpus by means of an interface that supports navigation through explorative inquiry and training in practice. To support learning in and with a corpus, the interface must emulate how learners explore the text for comprehension and then use the corpus to train in the essential skills of language learning. Using the encoding and linguistic annotation of the ETCBC corpus, Bible OL therefore offers a choice between a display of texts and a selection of exercises. Furthermore, when learners are logged in with a Facebook or Google account, they will have access to statistics on their ongoing learning outcomes.

The new way of language learning breaks away from the older grammar-translation method which depended both on the teacher introducing subjects to be memorized and offering some practice drills on selected forms and texts, and on the students reproducing in class or in tests the information the teacher had previously provided, as described in [Winther-Nielsen, MS]. The new corpus-driven architecture is designed for a different learning experience, implementing the principle of persuasive learning for continuous integration of inquiry and practice through seamless movements back and forth between displays and exercises.

---

[12] PLOTLearner can still be downloaded in the format delivered to the EU project at [http4]. In early 2015 we updated the program with the new open access ETCBC database, but the PC program will not be developed further; instead all new development will be invested in the online technology.

[13] These open resources are accessible at [http2] and [http5]. After almost four decades, research on the database has since 2014 been licensed under a Creative Commons Attribution-NonCommercial 4.0 International License. The interlinking of the two interfaces means that students can click from the display of the corpus in one tool to a display of the same text in the other tool. Searches in SHEBANQ can be uploaded for learning practice in Bible OL.

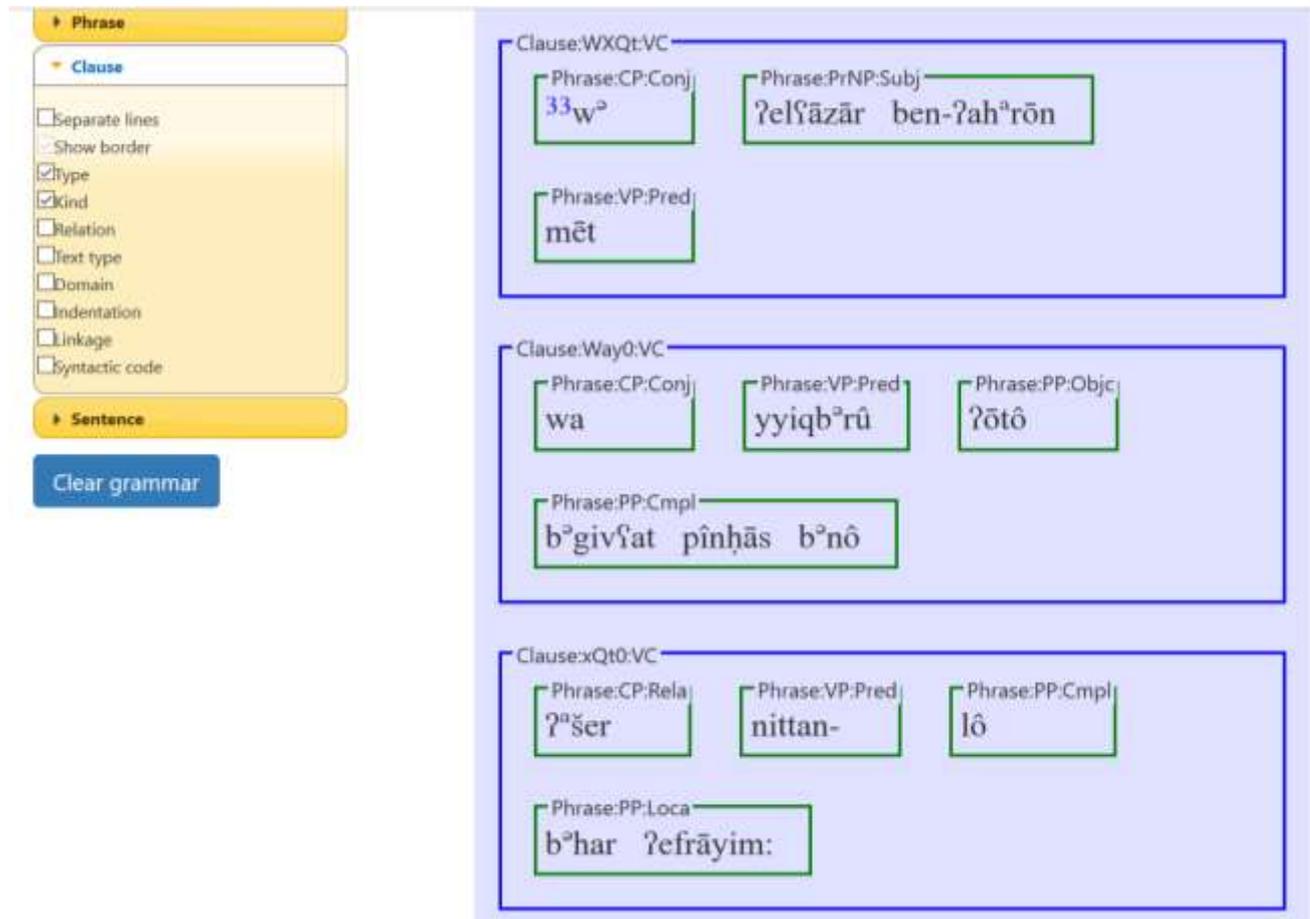

Figure 1. Joshua 24:33 in display from Bible OL (see discussion below and [http19])

This redefines the source of learning, the role of the learner, and the process of learning in three fundamentally new ways. First, the presentation of the content of the learning is moved from a teacher-defined lecture to an interactive corpus. The content of this corpus can be defined as a macro learning object in the sense that the database stores millions of potential learning objects that can be triggered through the interface. The essence of all instruction is to guide learners into an interactive flow by helping them to actively explore grammar through the corpus and to comprehend the categories of the grammar through the interface. For Hebrew words, the interface offers the choice of exploring information on the forms used in the texts, on lexical information, and on morphology. For phrases and clauses, it provides information on their structure and function as is illustrated by the text display in figure 1, later explained in detail in figure 2.[14]

---

[14] Other figures illustrating display and exercises are published in [Winther-Nielsen, 2014: 83]. In the left-hand navigation pane, the learner can select display of word, phrase, clause and sentence level information. A right-hand pop-up pane shows all content in the database when the cursor hovers over a word. The design is responsive, so on smaller displays the grammar selection and navigation is above the text, while the database pop-up shows in a new window in front of the text.

Second, all active learning is directed by the learners' interests. This is a major contrast to the traditional study of Biblical Hebrew, which often deprives students of any motivating curiosity due to the time they initially have to spend on learning to read; often students only get to read and interpret a text in a following term. In Bible OL, not only is the corpus content available in full display to be explored at all levels, but new learners can start with the text in transliteration to study the constituents of the phrases and clauses illustrated in figure 1.[15] The transliteration not only helps linguists or laymen with no prior training in Hebrew to be able to read the Hebrew texts, but first and foremost it supports interesting task-based language learning activities right from the start by engaging with the corpus. We are convinced that there is no stronger motivating force than the ability to wrestle with interpretations of the Hebrew Bible from the start.

Third, the most essential component in this new approach to corpus-driven learning is the opportunity of continuous practice. Corrective feedback came sparse and late in the old translation method, and it is unreliable in an immersion method, but here the instant feedback from an authentic and authoritative source is the main formative process. Learners can practice forms and train skills for analysis as long they want, to the depth they desire, and for the tasks they choose. No language skill is acquired without extended training for months. During self-directed training, learners can explore the text again and then continue to practice, or use some of the additional teaching material and then resume. Instant and reliable feedback motivates the learner to engage with the text and then to test and solidify new insights through practice. To support self-directed learning, PLOTLearner offered statistics that could be downloaded and emailed to the teacher, called the facilitator in this system, since he or she performs various roles such as instructor, supervisor, or designer of learning. Now under development for Bible OL the learner can visualize his or her own progress on a graph projected onto a view of the mountains from Mount Sinai. A graph shows the learning progress as a pathway into the mountains where the first alphabet was used by Semitic workers in the mines on the Sinai Peninsula some 3,500 years ago.[16]

These three components – the corpus-driven flow, the learner-directed force, and the task-based practice – can be contextualized by the facilitator. Teachers, instructors, or advanced students can scaffold the corpus with other kinds of learning objects like textbooks, pdf-documents, or videos. The project uses Moodle, the world's most popular open and free learning-management system, to store learning objects from courses, but any system or web page can be used. The system can also automatically scaffold from a picture database for illustrating a virtual word around the text explored by the learner, if the metadata of the picture has references to the text under scrutiny.[17] The interface is therefore not only learner-friendly, but also teacher-effective, enabling facilitators to improve on their teaching and achieve more efficient outcomes.[18] We now know that one reason why this new technology is not catching on worldwide is that many language teachers resist technology. The tool must, therefore, be able to persuade teachers that they gain from this new kind of textual interactivity, because they can adapt it to their true needs. Also, colleagues can join us and contribute to the development of new content as well as implement their own solutions. Dr. Oliver Glanz from Andrews University in particular has helped us improve the glosses and has provided labels for verb classes, so that Bible OL is currently the only tool in the world that can assist students in practicing irregular Hebrew verbs in the context of the biblical texts. In Madagascar, teachers are now using the system in rural areas without internet access. For other languages, we have translations of the interface into Spanish, Portuguese, and Chinese to facilitate learning and the creation of learning objects in other languages. These are just some examples of how our learning environment can help teachers join forces globally to develop a new kind of intertextuality for the learners of their own classrooms.

---

[15] Teachers subscribing to the Communicative Language Teaching Method often reject the use of transliteration and instead focus on speaking modern Hebrew detached from the ancient text. Our approach in contrast stresses authentic texts. We could provide sound files for the corpus, but we believe that providing a transliteration of the Hebrew text according to the reading of the text by Israeli speakers supports the development of pronunciation.

[16] See the manual at [http7] as well as the short discussion in [Winther-Nielsen, 2013b: 58].

[17] The picture database was developed in EuroPLOT, see [http6] and the scaffolding environment described in [Winther-Nielsen, 2013a: 27-28]. A new project will improve on the database and the scaffolded content.

[18] In the view of [Winther-Nielsen, 2014: 90], the "tool is unique in supporting this almost unlimited repurposing on any topic covered by one of the World's best databases for the Hebrew Bible".

In sum, Bible OL is a tool that invites learners to engage with the corpus, and it can be used long before they are able to read, parse, and analyse the Hebrew texts and even before they attend a Hebrew class. Once they engage, Bible OL offers the force and flow needed for students to learn the language. The tool can also help teachers solve new tasks, and it can inspire learners to a lifelong study of the Hebrew Bible for their professional lives. In this way, the interface implements a new persuasive model for interactive learning from corpora very well.

**1.3 Tasks in the New Global Learning Environment**
Computer-aided processing of interactivity is a pedagogical model for a tool and a basic principle in our design for persuasive learning, as well as being something that must happen in real life. In this sense, the success of any software and the models embedded in it are at best grand concepts, unless they are implemented and able to do the task envisioned by designers of learning.

Countless different factors can slow down and impede a successful implementation: we have run into many unexpected obstacles. Our implementation is still at a small scale, but fortunately we are able to demonstrate how the system works. The tasks supported by the interface are the basic tasks in any language-learning classroom, such as vocabulary learning, the parsing of morphology, and the syntactic analysis of phrase, clause, and text structure. As described in [Winther-Nielsen, MS]:

> The interface offers a translation informant for displaying and checking of glosses to assist memorization of vocabulary in context. The strength of this feature is that glosses can be selected according to frequency of occurrence for exercises, and thus improve on text-driven vocabulary acquisition. As a typist helper, the interface will check the learner's writing, reading, and spelling skills. Typing Hebrew text helps learners observe the details in the foreign script and scrutinize the exact visual shape of the writing system. The most important function is the syntax visualization which shows the hierarchical layers of words, phrases, clauses and sentences in the text and then allows for practice of this knowledge.

In the EU project, we gathered and analysed considerable quantitative and qualitative data from learners who tested PLOTLearner during the prototyping and piloting of the tool in Denmark, Sweden, and Madagascar as part of our development of an effective persuasive technology. The best data came out of the Lutheran Graduate School of Theology (SALT) in Fianarantsoa, Madagascar, where PLOTLearner was implemented as a core learning technology in the teaching of the Hebrew Bible in the Malagassy Lutheran church.[19] After the EU project ended in 2013, we decided to move all repurposing of the technology and new projects into a newly founded Global Learning Initiative [http8]. Our goal is to implement open persuasive learning solutions around the globe, and especially in countries with limited financial and technological resources, but with a high demand for Biblical Studies. The goal of the new project is to disseminate the results obtained in Madagascar and provide support for the implementation of similar projects in Africa, Asia, and South America. Another major goal for us is to develop a similar solution for learning of New Testament Greek.[20]

---

[19] These data are presented in [Winther-Nielsen, 2013b: 58-59], with further references to the evaluation reports with all the details on the prototyping and piloting 2010-2013.
[20] Claus Tøndering had by late 2014 implemented the open source corpus of Nestle 1904, Dobson Glosses, and the syntactic analysis provided by Global Learning Initiative. Harold Kime provided his e-learning Moodle course for New

As part of this project, I taught a course at SALT in October 5-17, 2015. The goal was to assist the local associate professor, Olivier Randrianjaka, in taking ownership of the project and the technology. I was asked to teach all 120 students in the mornings. In the afternoons, I offered lab sessions for advanced students, supervising their practice. This kind of full schedule worked well as a facilitation of learner-directed practice. All 120 students were introduced to my new guidelines for how to parse a Hebrew verb, but the level of the students' knowledge of Hebrew and English varied enormously. Assisted by my Malagassy colleague, I used Bible OL to support collaborative learning through student group discussions, promoting fun and competition through persuasive learning. In a flipped classroom environment, we had good results from turning learners into peer-instructors and teachers into facilitators. Corpus-generated exercises were projected on the wall of the lecture hall and were discussed in the groups and then commented on by students. This kind of co-teaching for interactive learning is an advantage for supervision of corpus-driven learning for large groups of diverse learners in global settings. It promotes co-teaching and student-directed collaboration fostered by the corpus. The tasks delegated to the groups could be used for training of teaching assistants.[21] My vision is that learner groups are supervised by advanced students, but this kind of intertextuality requires a fundamental departure from traditional education.

| Filename | Start at | Duration (min:sec) | Seconds per right | Cor-rect | Wrong | Correct per minute | Accu-racy | Pro-ficiency |
|---|---|---|---|---|---|---|---|---|
| Vocabulary 281-300.3et | 2016-03-01  13:39 | 00:46 | 9.2 | 5 | 0 | 1.3 | 5 | 1.3 |
| Vocabulary 281-300.3et | 2016-03-01  13:37 | 00:51 | 12.8 | 4 | 1 | 0.94 | 5 | 0.75 |
| Vocabulary 281-300.3et | 2016-03-01  13:36 | 00:57 | 14.3 | 4 | 1 | 0.84 | 5 | 0.67 |
| Vocabulary 281-300.3et | 2016-03-01  13:35 | 00:41 | 8.2 | 5 | 0 | 1.46 | 5 | 1.46 |
| Vocabulary 281-300.3et | 2016-03-01  13:34 | 00:32 | 8 | 4 | 1 | 1.5 | 5 | 1.2 |
| Vocabulary 281-300.3et | 2016-03-01  13:32 | 01:05 | 13 | 5 | 0 | 0.92 | 5 | 0.92 |

Table 2. Logbook for learner

For her dissertation, my PhD student Judith Gottschalk is working on ways to collect statistical data on learning results. All practice scores are logged as big data and can then be displayed in various ways to plot the performance of students and optimize learning experience. In [Gottschalk and Winther-Nielsen, 2013] we presented a pilot version of "Learning Journey", which Judith Gottschalk now uses for experimentation and testing, under the supervision by Claus Tøndering. Table 2 shows what kind of information is available for a particular type of exercise (Filename), performed at a certain time (Start at), and how long time it took in minutes:seconds (Duration(min:sec)). The

---

Testament Greek and offered to serve as facilitator online and overseas. Judith Gottschalk is the primary researcher working on developing the corpus-based learning of New Testament Greek for Global Learning Initiative, and Jean de Dieu is responsible for Greek in Madagascar.

[21] Videos document the fun, motivation, and ability for skill acquisition among students with little or no previous knowledge of Hebrew grammar. I expect that the learner-centred teaching approach will help many students to learn better, but a two-week course cannot work miracles.

information on how many seconds it took to produce a right answer (Seconds per right) is a helpful indicator of speed, and it goes without saying that it is also relevant to see the exact number of correct answers (Correct) and mistakes (Wrong). We also experimented with accuracy and proficiency, but these numbers are less easy to use.[22] The table finally shows how a learner can perform the same exercise several times, in order to finally get it right. Proficiency is clearly highest when he gets all five questions right in only 41 seconds.

|  | **Test October 15 2015** | | | | | | **October 24 2015 - March 3 2016** | | | |
|---|---|---|---|---|---|---|---|---|---|---|
|  | Seconds per right | Right | Wrong | Right per Minute | Accuracy | Proficieny | Ranks | Total Point | Time |  |
| *N01* | 6,4 | 212 | 28 | 0,04 | 8,57 | 0,03 | **3** | 3420.65 | 62:42:17 | 5 hours |
| *N07* | 9,2 | 141 | 27 | 0,04 | 6,22 | 0,03 | 24 | 195.76 | 12:43:24 | 1 hour |
| *N05A* | 9,4 | 159 | 33 | 0,03 | 5,82 | 0,03 | 13 | 387.03 | 17:30:25 | 15 minutes |
| *N10A* | 10,1 | 145 | 59 | 0,03 | 3,46 | 0,02 | 2 | 11011.89 | 116:12:54 | Many hours |
| *N06A* | 10,6 | 136 | 44 | 0,03 | 4,09 | 0,02 | 42 | 97.85 | 14:15:58 | 5 hours |
| *N47* | 10,8 | 150 | 36 | 0,03 | 5,17 | 0,02 | 5 | 2410.53 | 26:00:50 | Many hours |
| *N67A* | 12,3 | 128 | 70 | 0,02 | 2,83 | 0,02 | 79 | 36.76 | 06:35:13 | None |
| *N02* | 12,4 | 137 | 37 | 0,03 | 4,7 | 0,02 | 20 | 281.4 | 06:06:07 | None |
| *N09* | 14,2 | 124 | 26 | 0,03 | 5,77 | 0,02 | 45 | 146.46 | 04:07:43 | None |
| *N04* | 14,7 | 110 | 40 | 0,03 | 3,75 | 0,02 | 1 | 20601.36 | 64:11:21 | Many hours |
| *N49* | 30.6 | 58 | 32 | 0,02 | 2,81 | 0,01 | 226 | 0.14 | 00:15:07 | Little |
| *Nxx* |  |  |  |  |  |  | 56 | 71.44 | 03:54:17 | New |

Table 2. Progress from test in 2015 to 2016.

"Learning Journey" allows the facilitator to monitor the progress of learners. Some of the best students volunteered to do a final competition as a test on October 15, 2015, and they were asked to parse verbs for 20 to 30 minutes. This competition produced helpful results which were easy to rank. The results of the six best students would qualify them to become assistant teachers, according to the data in table 2. One was exceptional, achieving a score of 6,4 seconds per right answer, five were excellent at 9-11 seconds per right answer, and four were good at 12-15 seconds per right answer. I checked the data for all learners from SALT four and half months later, measuring the persistence and progress between October 24, 2015, and March 3, 2016. The exceptional student (N01) had practiced only for 5 hours, but had only dropped to ranking third among all users in the world, measured according to total points. The second best (N07) had practiced only for an hour and had sunk down to ranking 24 among all students in the world. This was in contrast to the student who had ranked 4 (N10). He had been away for part of the course, but had in the meantime practiced for 116 hours; his score raised to being the second best ranked among students globally. The most surprising result, however, is that the person who was number 10 in the test (N04), and was clearly at the bottom, had used the intervening months well and had practiced for some 54 hours altogether; he went straight to the top as the best globally. Table 2 therefore exemplifies the ability of corpus-driven technology to take a user all the way to the top, provided that he practices with Bible OL extensively and becomes almost "fluent" in his internalization of the corpus. In contrast, four other students who were in the lower half in the test did not practice much afterwards and they now rank lower in comparison to other learners in the world. This system in fact allows not only learners, but also facilitators to direct the students' practice since all learning processes are plotted.

---

[22] In [Gottschalk, 2014: 17], the degree of automatization is measured as proficiency, calculated as "sum of right answers / right answers per minute." Accuracy is an attempt to measure to what extent some students achieve a high speed, but also submit a high number of wrong answers; this is calculated as "(sum of right answers + sum of wrong answers) / sum of wrong answers", i.e., it measures the degree to which learners have responded correctly in comparison to wrong answers. Further experiment will show how much statistical data facilitators and learners need, and especially how this can be visualized in a persuasive manner.

My own class of thirteen students at Fjellhaug International University College Denmark in the fall of 2016 provided more fresh learning data. An intensive course of 20 ECTS credits in Biblical Hebrew equals more than 500 hours of student work. After 4 months of study, there is a final test on translation, morphology, and syntax covering 30 pages of the Hebrew Bible, in accordance with the standards set for Biblical Hebrew in Norway. I am developing an effective corpus-driven flipped classroom. Bible OL offers the students instant feedback and they are able to monitor their own assessment around the clock, wherever they are. Students use my videos and documents to prepare for countless exercises that take them through the course material bit for bit. Satisfaction derives from observing their own improvement from a poor grade all the way up to a B or A. They are required to do 100 hours of practice, documented in Learning Journey, as well as to pass two tests, one for beginners on Gen 1:1-5 and one on the basics of Hebrew Grammar covering Genesis 1-3. Seated together, the students work on exercises individually, but discuss their tasks with one another. This learning environment increases peer instruction and social learning. They only attend class when they need supervision and even part of this is done online. Hence, the teacher in this kind of flipped classroom is the Bible OL, which has the role of tutor and trainer, while accompanying videos and texts enhance pedagogical instruction. The human teacher serves as a facilitator who designs the digital curriculum for learning, supervises the learning progress, identifies the mistakes of each individual learner, and grades the result. Students and their facilitators are guided along on their learning journey in class encounters and individual sessions.

Table 3 Weekly Results in FIUC-Dk class from September 5 to October 13 2016

| | 5.9 | | 13.9 | | 19.9 | | 26.9 | | 3.10 | | 10.10 | | 13.10 |
|---|---|---|---|---|---|---|---|---|---|---|---|---|---|
| 1. | C+ | 37.39.59 | C+ | 47:39:31 | B- | 57:05:23 | B- | 66:03:25 | B- | 74:05:08 | C+ | 91:01:44 | C+ | 100:18:45 |
| 2. | B+ | 18.11.16 | A- | 29:11:20 | A- | 35:04:59 | A- | 48:00:13 | A- | 59:20:55 | B | 71:29:27 | B | 80:22:45 |
| 3. | C+ | 14.51.02 | C | 24:27:12 | C+ | 26:25:40 | C+ | 41:31:04 | B- | 67:54:20 | B- | 94:49:19 | C+ | 100:11:10 |
| 4. | B+ | 17.35.45 | B+ | 32:12:20 | B+ | 47:35:48 | B+ | 64:48:08 | B | 78:08:16 | C+ | 99:18:15 | B- | 102:34:56 |
| 5. | C+ | 12.07.37 | B- | 22:19:24 | C+ | 36:11:03 | C+ | 61:31:18 | B- | 93:12:08 | B- | 98:56:14 | B- | 100:36:28 |
| 6. | A- | 29.17.40 | A | 36:53:36 | A- | 43:35:16 | A- | 52:22:30 | A- | 62:50:06 | B+ | 72:30:03 | B+ | 81:40:46 |
| 7. | D | 36.00.00 | C- | 46:40:28 | C | 61:09:50 | C | 71:23:52 | C | 86:25:54 | C- | 93:37:10 | C- | 101:18:50 |
| 8. | B- | 19.16.14 | B | 26:59:18 | B | 34:19:57 | B | 45:07:10 | B | 51:36:36 | C+ | 70:56:23 | C | 82:17:19 |
| 9. | B+ | 14.01.26 | B+ | 26:25:25 | B | 32:06:02 | B | 42:44:31 | B | 52:38:35 | C | 69:53:38 | C | 80:06:33 |
| 10. | B | 07.28.38 | A- | 20:23:53 | A- | 29:58:08 | A | 52:54:31 | A | 71:49:44 | A | 90:26:23 | A | 102:28:28 |
| 11. | C- | 25.19.47 | C | 34:26:39 | C | 40:14:18 | C | 50:53:52 | C | 63:57:27 | D+ | 95:35:04 | C- | 100:11:30 |
| 12. | B | 18.55.59 | B | 27:46:59 | B | 30:55:03 | B+ | 45:02:14 | B | 57:19:46 | C+ | 70:44:04 | C+ | 80:22:38 |
| 13. | B | 36:46:45 | B+ | 54:01:09 | B+ | 59:28:39 | B+ | 69:00:03 | B+ | 80:17:23 | D+ | 92:41:23 | D+ | 100:13:22 |

The tables illustrates the kind of data a facilitator can extract from Bible OL's "Learning Journey". Table 3 illustrates the record of grades and the total time of exercise practice. Learners can monitor their progress and decide their own goals for each individual task. When they finish an exercise, they can click the "SAVE outcome" button in order to observe their progress on a certain skill, and when they are satisfied, they can click the "GRADE task" button in order to share this result with their facilitator. Facilitators can monitor these as often as they wish and use the results into other programs or documents. The students can always improve their grades by redoing an exercise. As a mid-term requirement, the students in Copenhagen had to log a total of 100 hours of practice by October 13. It was then possible for the facilitator to observe how their grades improved (bold), or worsened (underlined), for whatever reason.

Table 4 illustrates how the teacher can track the performance of a learner on a specific task like translation and vocabulary learning.

Table 4. Student 3 Performance on vocabulary (English) and verb parsing (verb-class)

| | | | |
|---|---|---|---|
| English | 3357 | 81.14 | B- |
| Verb_class | 56 | 35.71 | F |

Table 5 illustrates how the teacher can document the grade of test.

Table 5. Student 3 Performance on second test

| | | | |
|---|---|---|---|
| Test2A_Part Of Speech-5 Questions | 70 | 88.57 | B+ |
| Test2B_Nouns-10 Questions | 48 | 93.75 | A |
| Test2C_RegularVerb-10 Questions | 50 | 80 | B- |
| Test2D_Irregular Verb-10 Questions | 72 | 91.67 | A- |
| Test2E_Translation-5 Questions | 22 | 68.18 | D+ |

The learning journey for each individual student reflects the individual leaning styles, goals, and motivation. The grades cannot be compared, because they reflect how far learners have progressed (level of difficulty), and whether they used helps or tested themselves by memory (number of mistakes). However, the final exam results for this class were promising. Even though two students had to retake the exam, in the end everyone passed, half with high marks and half with low. Thanks to an average of 600 hours of work in the class in Copenhagen, I had invaluable feedback for the improvement of Bible OL, exercises, and learner resources. We have data to refine learner data and improve the Learning Journey with graphical visualization, badges, and other helpful features. What we have at the present stage of development works well and will allow us to improve for new learners in the future.

Even at this seminal stage, these two projects, one in a Majority World setting in Madagascar and one in a European setting in Copenhagen, illustrate the effectiveness of using a corpus-driven flipped classroom. I have had good results from turning learners into peer-instructors, but less success in recruiting teachers trained in the traditional roles to become facilitators. The new environment for collaboration is successful thanks to the corpus-driven instant feedback and continuous assessment for monitoring of learning outcomes and for surveillance of outcomes and grades. Although "Learning Journey" is still a prototype version, it is used by facilitators at [http9]. The developer and coordinator of "Learning Journey", Judith Gottschalk, plans to reprogram Learning Journey and test it on learners of Greek and Hebrew in 2017 and onwards.[23] Currently we are looking at how facilitators can use the data and how "Learning Journey" could be optimized. All wrong answers are registered for inspection and analysis; these can help facilitators improve their instruction and differentiate their teaching. Data like those harvested in Madagascar can help facilitators keep track of scores and plot the outcomes of ongoing learner practice in class or online; results can be measured in relation to global outcomes.[24] Tests, user-interests, and resources will determine how far a new corpus-based paradigm for data mining and interactivity can support individualized journeys enhancing performance and persuasion though a corpus.

---

[23] The goal is for Judith Gottschalk to integrate "Learning Journey" into Bible OL with menu points giving access to (1) International Ranking, (2) Logbook, (3) Learning Graph, (4) Badges (personal communication in document for her doctoral dissertation).

We envision offering this tool to many more global users and giving access to thousands of learners. We hope in this way to be able to build better profiles of learners and to be able to predict persuasive force and flow from global data. As we go ahead, hopefully the tasks and the feedback from users will help us formulate better models and improve the technology. Because Bible OL is completely without license restrictions, we hope that programmers will join us in designing better solutions, and that more global learners will use the interface in their mother-tongue. We hope to build a global community of facilitators, and will look for funding for new projects in Africa, Asia, and Latin America.

# II TOOLS FOR TEXTUAL CRITICISM

From this new kind of intertextuality in learning, defined by learners actively engaging in inquiry and practice with, in, and around the corpus, I move on to a different kind of tool for textual criticism and describe another type of task that plays a role in a computer-assisted processing of texts from the Ancient Middle East. Emanuel Tov, a world expert on the textual criticism of the Hebrew Bible, has recently completely revised and expanded his textbook, *The Text-Critical Use of the Septuagint in Biblical Research,* now published in its third edition in 2015. He explains that a revision was warranted not only by new developments in the study of the Hebrew manuscripts from the Judean Dessert and a better understanding of the Greek translation, but also by "the computerized approach" [Tov, 2015: xi], the topic of discussion here.

If we had linguistically annotated open access corpora similar to the ETCBC for all resources of textual criticism, we would be able to support a new kind of textual criticism. Here we aim to describe which tools learners need for specific tasks, and which resources are already available, first in general, then for Hebrew manuscripts, and, finally, for translations of the Hebrew Bible. I will discuss the Logos Bible software as one of the scholarly tools that offer the best information architecture and the widest range of technological support for learners studying the original languages, giving access to specialist resources in a user-friendly interface. I also include references to similar resources in the Accordance program. This discussion will prepare for the discussion of corpora for textual criticism in the third and final section.

**2.1 Resources for Textual Criticism**
Most students and teachers at universities these days receive little fundamental training in philology and textual criticism, though in the past that was part of traditional elitist university training. Whether we like it or not, the modern curriculum must adapt its training in textual criticism to a different pedagogical and practical approach, if indeed the Biblical languages are still taught at all. At the core of my proposal for teaching textual criticism is the conviction that not only are the resources available that are crucial to the study of the ancient texts, but that they also can foster skills needed in the knowledge workers' workplace. Because textual criticism is not supported by Bible OL and SHEBANQ, the goal here is to describe how open corpora can be integrated with tools currently available for the study of other textual resources bearing on the Hebrew Bible.

---

[24] "Learning Journey" is not yet a full-fledged intelligent tutoring system for machine-learning; for now we use the statistics for experimentation and development. In the dissertation project, Judith Gottschalk plans to explore support for gamification and collaboration.

Resources for textual criticism are primarily still published commercially as printed books. These are less interesting to Western students oriented primarily or solely to digital media, and are unavailable to global learners who cannot afford copyrighted material.[25] The standard scholarly treatment on textual criticism by Emanuel Tov, the *Textual Criticism of the Hebrew Bible,* was first published in 1992, appeared in a second version in 2001, and is now available in a third revised and expanded edition [Tov, 2012]. This technical work cannot be used readily by the average student; to use it, the Hebrew language and exegesis curriculum will need pedagogical introductions like the complete revisions of [Würthwein and Fischer, 2009/2014] or [Brotzman and Tully, 2016].[26]

Of the many projects on textual criticism in the past, two currently stand out as promising for the future.[27] The first is the *Textual History of the Bible* (THB), for which [Lange, 2016: 1] claims that it offers "several noticeable paradigm shifts in the field of text criticism", chiefly because textual witnesses are now "studied as texts and traditions in their own right", instead of as isolated variants. However, the project also amply illustrates the problem of the increasing costs of commercial corpora, which is bound to reduce their accessibility even for students in the West.[28] The THB furthermore raises another issue by taking a plurality of biblical texts for granted. For this reason, scholars no longer attempt to reconstruct a "supposed biblical *Urtext*, but aim as much to reconstruct the entire textual histories of the biblical texts." In this sense, the THB blurs the old distinction between Higher and Lower Textual Criticism, and textual criticism turns into a redaction history of the Hebrew Bible.

The second and apparently even larger project, *The Hebrew Bible: Critical Edition* (HBCE) aims to produce an entirely new text of the Hebrew Bible based on modern scholarship with a critical text and extensive text-critical introduction and commentary. It is probably going to be even more inaccessible than the THB to the average student.[29]

Both of these projects illustrate two trends and the challenges facing the student of textual criticism of tomorrow. The first problem is that these new online resources are not moving in the direction of open corpus-driven data, and learners cannot expect to be able to use any of those online resources. A more important problem is that the paradigm shift to redaction criticism in the THB is a belief to be tested in terms of a choice between diachronic and literary readings of the Hebrew Bible. This shift will be even more far-reaching when we get an entirely new hypothetical *Urtext* in the HBCE. These trends inevitably raise the question of the scholars' assumptions and approaches to the texts.

If textual criticism turns into redaction criticism, the notion of a stable canonical corpus vanishes. How to solve this challenge will be one of the tasks that needs to be addressed in a proposal for a new textual corpus criticism, and it can only be meaningfully discussed within the broader context of exegesis and in concrete corpus-driven text-analysis, as we will illustrate for Joshua in the end (section 3, sub-section 3). This is an important issue for the intertextuality of the ancient texts.

**2.2 Tools for the Study of Hebrew Manuscripts**

---

[25] In due time hopefully projects will offer open access to high-quality scholarly corpora online following the new trend of copyright-free access to books and resources in the humanities.

[26] For other introductions on textual criticism, see [Tov, 2015: 1-2], [Wolters, 1999: 19-20, n 1], and [Wegner, 2006].

[27] For other projects and resources see [Tov, 2003], and for the use of computers in general, see [Tov, 2008].

[28] In 2015, Brill set the price for access to the online version at 2,700 Euro, see the Brill site at [http10]

[29] The project was the Oxford Hebrew Bible, and [Heidel, 2013] is the general editor of the critical Bible. Each book will be published in a separate volume, except for single volumes of Minor Prophets, Megillot, and Ezra-Nehemiah. See further information at [http11]. The price of so many specialist volumes will be exorbitant.

Given the challenge of the costs of printed and even online resources, my main interest is to address the issues raised by [Tov, 2011] on the pros and cons of electronic tools. I will proceed in an eclectic manner, based on my own experience with digital resources for the PC. I have been a dedicated user of Logos Bible Software since 2003, and especially of the resources published by the German Bible Society. I leave it to others to extoll the merits of Accordance, Bible Works, and other popular resources.[30] I will first address the resources for the study of the manuscripts in Hebrew, namely, the Hebrew Masoretic Text (MT), Dead Sea Scrolls (DSS), and Samaritan Pentateuch (SP).

The primary aim of textual criticism is to help learners work with the earliest fully extant manuscript of the Hebrew Bible, the Codex Leningradiensis from 1008 AD, which is available in the printed editions of the *Biblia Hebraica Stuttgartensia* (BHS) from 1977, and now printed in its fifth edition as [Elliger, Rudolph and Schenker, 1997]. Learners have open access to this text in Bible OL and SHEBANQ. Even if this edition is the sole scientific edition available, [Tov, 2011: 248] rejects its apparatus "as unsatisfactory for text-critical analysis since it provides far too little information and is much too subjective." However, not only is the printed scholarly BHS edition a reliable reproduction of our primary Hebrew manuscript, but it also contains a critical apparatus that provides manageable information to get a student started in textual criticism without too many scholarly filters. The BHS gives access to the MT, so named after the Masoretes, or tradents, who studied, copied, and preserved the Hebrew text between 500 and 1100 AD, and gave us our present text in Tiberian Hebrew.[31] Adolf Schenker initiated a new project for the German Bible Society in 2004, the *Biblia Hebraica Quinta* (BHQ), which will give access to earlier pre-Tiberian witnesses and include references to the famous Aleppo Codex rescued out of a fire in 1947, as well as the Yemenite codex of the Writings, Cambridge Add. Ms. 1753, and fragments found in the Qumran caves and in the Judean Desert. This edition is clearly an improvement and will no doubt be the best option for advanced students in the foreseeable future, even if this edition also is rejected by Tov as "merely a selection of textual data" [ibid].[32]

Computer-literate students have a helpful solution for accessing the BHS thanks to the important contribution of the German Bible Society. From 2004 to 2012, it marketed three versions of *Stuttgart Electronic Bible Study* (SESB) with an introduction by [Hardmeier, Talstra, and Salzman, 2009]. Logos Bible Software now offers some of the tools in a small product containing the indispensable critical apparatus of the SESB edition alongside the apparatus for the Greek New Testament UBS edition.[33] This interactive access to the study of the BHS with its text critical apparatus provides learners a viable alternative to the costly and unwieldy printed editions. Students can also benefit from a digital integration of other scholarly resources for textual criticism from Logos in a state-of-the-art user interface that greatly enhances intertextuality for the Hebrew Bible. Last, but not least, it uses the ETCBC database and therefore is a helpful expansion to learning and research through open access resources like Bible OL and SHEBANQ.

The challenge for the student of textual criticism is, of course, that not only do we not have the first manuscripts of the original authors (the *autographs*), but the gap between the original compositions and the final editions may span up to two thousand years or more. Because our scholarly BHS/BHQ editions of the codices date from the tenth century AD and onward, our problem is how to bridge the gap back to the earlier and potentially more reliable evidence on the text, and how to make this evidence accessible.

---

[30] [Tov, 2015: 35] lists products which are by now out of the market (SESB) or of marginal relevance (e.g., Gramcord and WordSearch); see further references on [ibid: 37] and his excursus on CATSS [ibid: 110-111]. For a review, see [htpp12].

[31] Hence [Anstey, 2006] more precisely establishes the phonology and grammar of Tiberian Hebrew.

[32] Several facsimile editions have been published, see [http13]. The goal is to complete the BHQ by 2020, one hundred years after the first edition was published by Rudolph Kittel, see [http14].

[33] The Core Bundle of the Stuttgart Scholarly Editions would be competitive if it not only offered the SESB 2.0 with Apparatus, WIVU Introduction and Constituency Trees, but also the Nestle Aland 28 with apparatus, but the latter is missing, see [http15]. Logos' Academic Discount Programs minimizes the expenses for students at western institutions, but outside the West students can only acquire these tools through donations and subsidized resources.

The task of textual criticism initially is to evaluate the reading variants that the experts have assembled and classified; however, this evidence mostly confirms the dominance of the MT tradition from around 100 AD. The real challenge of textual criticism starts with the earlier Hebrew manuscript evidence emerging from the Judean Desert dating from 250 BC to 135 AD, found primarily during the period 1947-1961. Among the DSS manuscripts were some 200 samples of the Hebrew Bible. After some initial delay virtually all were published by the mid-1990s. The standard study edition by [Martinez and Tigchelaar, 1998] is available from Logos, but the translation and comments on the scrolls are not in (see [Abegg, et al., 1999]).[34] The evidence from these finds varies greatly, and the early excitement over the confirmation of the MT is long gone: some manuscripts support the MT, while others reflect the SP, or other ancient translations. The Hebrew text of the SP is in general considered to be a sectarian expansion in support of Samaritan theology, yet eight of the DSS scrolls arguably support an earlier Pre-Samaritan text.[35]

This kind of evaluation of the Hebrew witnesses is a second task that must be integrated into a new textual corpus criticism. The question is to what extent deviations in the Hebrew texts are evidence of revisions in earlier stages of the text, or whether alternative theories on the formation of the text would explain these data better.

**2.3 Tools for the Study of the Translations**
For the study of the text of the Hebrew Bible we also rely on ancient translations of it: the Greek Septuagint (LXX), the Syriac Peshitta (S), the Latin Vulgate (V), and the Aramaic Targums (T). Studying an ancient text through the lens of translations is an indirect approach, which represents a kind of mediated textuality for ancient texts. The many translations testify to the fact that the Hebrew Bible has had a special status as a canonical and holy text through time and in different cultures.

The evidence from translations is diverse. The LXX is in itself a challenging version. The most deviation from the Hebrew text can be found in the Greek translation of the book of Jeremiah, which is roughly one-seventh shorter than the MT. Two fragmentary Hebrew DSS manuscripts, the 4QJer$^{b,d}$, confirm the shorter text of the LXX against the MT, and other scrolls confirm details of the LXX. This fact pushes evidence from the LXX version to the forefront of textual criticism. As for resources, the standard study edition of [Ralph and Hanhart, 2005] is available in Logos, but not the English translation of [Pietersma and Wright, 2007]. There are almost limitless scholarly resources for the study of the LXX,[36] but the most important are the 24 volumes of the *Göttingen Septuagint* 1931-2004, available through Logos.[37] It is also of great help that Logos has published Tov's *The Parallel Aligned Hebrew-Aramaic and Greek Texts of Jewish Scripture* (2003). This work is an interesting case of a resource that started as a database and never was printed, but is now available as an integrated digital resource. Other later and less challenging evidence can be found in the Syriac and Latin translations as well as in the Aramaic paraphrases.[38]

---

[34] For references to the DJD series, see [Wolters, 1999: 20 n. 2]

[35] The SP is reprinted in [Von Gall, 1993] and now translated in [Tsedaka and Sullivan, 2013], and recently discussed by [Anderson and Giles, 2013], but this is not available as a Logos resource.

[36] See most recently [Tov, 2015] for LXX lexica (33-34), grammar (34-35), translations (35), editions (35-36) and concordances (107-108).

[37] See the references to the publications at [http16]. It provides the most authoritative critical apparatuses to date and the volumes include evidence from contemporary Jewish and Christian sources. It was initiated by Ralph in the 1920s and has for the last decades been edited by scholars like John William Wever.

These translations define a third task for a new textual corpus criticism proposal: are translations free dynamic equivalent translations or do they testify to the transmission of a more original text? How can we know?

This overview of essential digital resources in projects, manuscripts, and translations only scratches the surface. Among the thousands of resources, learners with sufficient means will be able to use great scholarly tools through Logos for their education or research projects. What is not easily accessible as software will probably also be out of reach as printed versions. The disadvantage of commercial programs is that they are not available for multiple use in contrast to printed editions which can be used by many students in libraries and retain their antiquarian value as a commodity.[39] [Tov, 2011] may have a point in his critique of the BHS/Q, but this is the only commercial tool for all platforms, for better or worse. At least for any foreseeable future, commercial corpus applications are bound to set the standard for our tools and they will determine the goals for open, globally accessible, next generation tools for Biblical Studies.

## III. TASKS FOR TEXTUAL CORPUS CRITICISM

Thus far I have described digital intertextuality in terms of an open global access to interactive corpus-driven learning and in terms of useful commercial resources for textual criticism published by Logos and other software companies. In this third and final section I will define intertexuality from the perspective of tools and tasks that emerge from next generation corpus technology for learning and research on the Hebrew Bible. I will outline goals for a new textual corpus criticism based on digital technology developed along the lines and principles implemented in Bible OL and SHEBANQ on the basis of the ETCBC corpus, but expanded for corpus-driven textual criticism.

The Logos tools considered thus far must set the standards for designers of new open-learning software to meet the needs of the next generation of Biblical Studies; in turn these tools may inspire software companies. The tasks set for the new tools for the next generation textual corpus criticism involve handling Hebrew variants, Greek translations, and the earlier editorial stages of the texts, as well as determining what kind of corpus-based technology these tasks call for:

1. Can variant readings bear on the formation of the Hebrew Bible?
2. How can a Greek translation prove to represent a more original Hebrew text?
3. Should textual criticism be used for redaction criticism?

**3.1 Variant Readings in the Oral Transmission of Written Texts**
The first task for one using Bible OL, SHEBANQ, Logos, or any other interactive application or software is to learn how to evaluate variant readings from other Hebrew codices or scrolls. If all variants are not equal, how do we weigh them? Do they have any bearing on the formation of earlier stages of the texts?

---

[38] The Peshitta Institute in Leiden began in 1972 with the scientific publication of the Peshitta, and Logos offers these monographs by the [Peshitta Institute Leiden, 2006]. The Latin Vulgate is published as [Weber and Gryson, 2007] and also available from Logos. Finally, there are translations of the Targums by [McNamar, et al., 1987-1997]. Note that BHQ, LXX and V is included in the Stuttgart Scholarly Editions: Old and New Testament at [http17].
[39] The single-user restriction is imposed by publishers; perhaps rental software at [http18] will be an option in some pedagogical curricula.

Once facilitators have introduced the tools for textual criticism and the views expressed in the textbooks of the experts, learners start the real task of practicing the tasks of textual criticism. This is often a perplexing challenge and it is hard to decide between two or more different explanations for a phenomenon. When presented with a shorter variant in a particular case, the learner has an informed choice between "shorter is earlier" or "less is more", and how can we be sure?[40]

In this sense, textual criticism has to face a common trap for Biblical scholarship, pointed out by [Tov, 2012: 170], that "too often, scholars take abstract assumptions and preconceived ideas unrelated to the Scripture texts as their point of departure."[41] He assumes that the biblical books were written over many generations and underwent processes of revisions, in the case of Joshua-Kings, or different literary stages, in the case of Jeremiah and Ezekiel [ibid: 166]. In his view, the final copy is therefore probably preceded by literary crystallizations and this would allow for consecutive editions at the beginning. However, even if earlier compositional stages could not be eradicated, he still believes that "the original text(s) remains an evasive entity that cannot be reconstructed" [167].

The challenge posed by such assumptions are addressed in Carr's recent study *The Formation of the Hebrew Bible* [Carr, 2011]. His aim is to formulate a "methodological modest" form of transmission history with full control and repeatability according to standard empirical procedures.[42] At the core of the proposal of [Carr, 2011: 65] is an endeavour to pay "more attention to the tendencies of individual manuscript traditions and the ideological-theological and lexico-grammatical dynamics of each given case." The whole argument in Carr's new approach depends on the evidence for his assumption that the scribes employed a writing-supported oral memorization in their transmission of the Hebrew texts. The best explanation for the variants in the Hebrew manuscripts is "a mix of oral and written dynamics" [17]. He develops a tripartite characterization of variants. The first two kinds are aural variants caused by mistakes in hearing or dictation and graphical variants involving the confusion of letters or the skipping of lines. The third category are memory variants caused "when a tradent modifies elements of text in the process of citing or otherwise reproducing it from memory, altering elements of the text, yet producing a meaningful whole" [ibid]. These variants are not garbled due to mistakes made by scribes in copying, but "good variants" in the sense that both variants makes sense in their individual contexts.

Carr argues his case from Gilgamesh, the Temple Scroll, and dual traditions. The empirical evidence Carr brings up for these variants are variation of events in different order and different words [59-61]. They can involve word order shifts, semantic shifts involving lexical or synonymous variants, different designations for figures, and grammatical changes in prepositions and minor particles, in short "[s]maller scale shifts in a single word, particle, or grammatical expression" [62]. He further argues that shifts go in both directions and therefore are hardly evidence that one of the two texts was "systematically updated or otherwise revised a precursor" [ibid]. This empirical analysis of memory variants leads him to conclude that the transmission process was dominated by three broad tendencies:

---

[40] To illustrate, [Carr, 2011: 70-71], as many others, argue that the longer Hebrew text of Jeremiah, Samuel and the Proto-Samaritan Pentateuch must be later than the earlier variant texts in Hebrew and Greek. However, when Chronicles has minuses in comparison with Samuel and Kings, the shorter text is not explained as earlier, but as using other and shorter sources different from Kings [77].

[41] [Tov, 2012: 163] insists that scholars "cannot afford themselves the 'luxury' of not having an opinion on the original text of the Hebrew Scriptures". He first surveys two alternative models, but discharges the proposals by Kahle, Barthélemy, and Goshen-Gottsten on multiple pristine texts at the beginning [163-165], and instead argues for "an original text or a series of determinative (original) texts" [165; cf 167-169].

[42] [Carr, 2011: 36] rejects Biblical scholarship which tries to "reconstruct highly precise differentiations of potential precursor literary strata"; instead of hypothetical dependencies and secondary additions he wants to "see in our various editions of Hebrew Scriptural texts the distillate of a transmission-historical process" shaped by memory and performance [ibid].

the trend toward expansion [65], the tendency for additions to occur particularly at the beginning and the end [66], and the fact that usually eliminations involve contradictions [70]. The second trend was an abbreviation of parts of the tradition [88]. The third trend is harmonization or coordination, as he calls it [90].

Carr's model potentially offers a far more empirical basis for a corpus-driven study of variants. His point of departure is to a lesser degree unproven assumptions of hypothetical stages, postulated in order to explain the growth of the tradition.[43] Be that as it may, [Dershowitz, et al., 2015] discuss a computerized source criticism of Biblical texts, which we suppose could be used for collecting evidence for memory variants rather than putative sources, utilizing cutting-edge technologies in the field of computational linguistics. At SBL Atlanta in 2015, Joshua Berman and Moshe Koppel presented their Tiberias Project, a web-based tool for text categorization and authorship attribution of the Hebrew Scriptures. This tool may be available online for other researchers in 2018; it will enable scholars to conduct their own experiments on text categorization, using the ETCBC database. It remains to be seen through experiment whether this application can help us evaluate the sources and test the evidence for memory variants, but it does at least give us the ability to see far more features of a lexical and morphosyntactic nature emerge through text comparisons which will align the texts according to degrees of similarity.

Along these lines we can safely expect new solutions for computer-assisted processing of the intertextuality of the Hebrew Bible to offer new potential for the study of the transmission of the historical texts. The digital solutions will be less dependent on the literary presuppositions inherent in the older more deductive approaches in Biblical studies.

**3.2 The Data from Analysis of Translation Shifts**
The next task for a textual corpus criticism is to help students decide on how far a variant in the Greek translation is also a witness to a more original text in Hebrew. Do we just pick one blindfolded or is our preference based on our prior assumptions or best guesses? Is it possible to attain more solid ground for an evaluation of the value of translations?

Again Tov is our expert as well as our challenge, assuming that we can use the LXX as a tool in Biblical criticism.[44] According to [Tov, 2015: 236], the contribution of the LXX is clearly seen in the fact that what researchers used to see as scribal changes, glosses, and interpolations, are now interpreted as a part of the "presumed history of the biblical books and manuscripts", attesting to different stages in the literary development of the books.[45]

The challenge for a corpus-based approach is how to deal with retroversion of variants from Greek into Hebrew. Any difference from MT in a translation is not necessarily a variant, "because translators introduced many such details without relation to the Hebrew text before them" [ibid: 9]. The main criterion Tov uses in his attempt to reconstruct a *Hebrew Vorlage* is whether the Greek text is evidence of a literal or a free translation [18], assuming that a literal translation is also a witness to a more faithful translation [19]. In this sense it is a matter of grasping an individual translator's technique in order to define what is literal [21].[46]

---

[43] [Carr, 2011: 65] recommends the analysis of a "process that is betrayed by extensive verbatim agreement between traditions combined with occasional variation between expressions of similar or virtually identical semantic content."
[44] See [Tov, 2012: 37, n. 42, 38-39]. He wants to combine Lagarde's *Urtext* theory and Kahle's "multiple translations" theory and then go beyond to formulate a theory of "multiple textual traditions" [11].
[45] [Tov, 2015: 12] assumes that there was one Greek translation behind most of the LXX, but it was "not long preserved in a pure form". It split into several secondary textual traditions followed by a stabilization in 1st-2nd CE, but new textual groups emerging in Origenes and Lucian [11-12].

A new technological solution has been presented by the promising young co-author of the new second edition of [Brotzman and Tully, 2016]. The dissertation by [Tully, 2015] on the Peshitta version of Hosea gives him the necessary first-hand expertise for developing a model that can distinguish textual variants from translation variants. To overcome the challenge of retroversion, Tully uses modern theory on translations shifts, and from these he deduces the operational norms of the translator and ultimately the translator's overall approach. Modern translation studies help him identify shifts as formal correspondences rather than deducing them from putative sources. When norms are identified through categorizations of consistent patterns, the analysis based on translation theory will work its way back from data to norms and then to shifts, and in this way provide an informed analysis of textual criticism. Tully does not have a program or an application that will read the Hebrew and Greek texts and align them side by side, but he is advocating a methodology that works like a recursive computer program, in which later conclusions are fed back into earlier questions. If this kind of recursion were to be developed for computer-aided processing, the program could learn from previous solutions which would function as a substitution for the next case, rather than as a presupposition for the analysis. In the end, the program would gradually accumulate all decisions taken.

Tully's methodology illustrates what future projects should develop through machine learning and it clearly defines the second task for a new textual corpus criticism. Ideally, we should have a corpus of the LXX developed to the same scholarly level as the ETCBC database. We then could use an application for semi-automated analysis of translation shifts that measures correspondences and calculates regularities. Unfortunately, we are years away from a LXX developed to the level of the ETCBC database. However, a true corpus approach based on data patterns will eventually provide a more solid basis than dubious presuppositions on the growth of texts; it should proceed case by case rather than from one theory to another.

**3.3 The Task of Text Analysis and Joshua**
The final task emerging from the survey of resources concerns the question whether textual criticism should be turned into redaction criticism or kept completely separate from the issue of redaction. How do we ultimately use this evidence in our interpretations of the Hebrew Bible?

In [Winther-Nielsen, 1995], I worked with the ETCBC database for the book of Joshua. The evidence suggested to me that the MT in general is uncorrupted, with only minor scribal errors, in contrast to the deviations in the LXX which "mostly contains abbreviations and simplifications of the MT" [Winther-Nielsen, 1995: 21]. Nevertheless, a number of scholars and most recently and rigorously Graeme Auld, have claimed that the LXX is superior to the MT of Joshua.[47] However, I not only rejected the replacement of the MT for the analysis of the literary structure of Joshua, but also made the more fundamental claim for my discourse grammar that "it should not even attempt the common practice of reconstructing a more original text … prior to textual analysis" [ibid: 22].[48] Only when a competent Greek scholar has done a functional discourse grammar of the LXX text of Joshua can we compare and evaluate whether the Hebrew text should be rejected as inferior. In the ensuing discussion, Auld's position has not met with much approval.[49] However, the dissertation of [Den Braber, 2010] defends Auld's approach to Joshua along similar lines, even though she also endorses a grammatical analysis based on the ETCBC.[50] The price she pays is that there are no authentic data and only a community of faith which left "a tradition … embodied in subsequent revisions" [ibid: 236].

---

[46] [Tov, 2015: 22-25] defines five criteria for literal renderings: α internal consistency, β the representation of constituents of Hebrew words by separate Greek equivalents, γ word-order, δ quantitative representation, and ε linguistic adequacy of lexical choices.

[47] See the details in [Winther-Nielsen, 1995: 21, n. 32]: The main scholar in my discussion is Graeme Auld who believes that in some chapters the additions in the MT were as high as five times as many as additions in the LXX. Tov as well assumes that the LXX has important additions in 16:10, 19:47, 21:42 and 24:30, 33, and concludes that there existed a shorter *Vorlage* of Joshua in the third or second century BC.

[48] I was building on [Rabe, 1990] who calls for a synchronic textual criticism, and especially [Rabe, 1992: 292] claims that we need to do our analysis on a verifiable existing text analysed as a material and literary unit, only removing scribal errors.

We have argued with Carr, that Biblical scholarship should work with documented trends in the transmission history of the text. I will therefore test the case [Tov, 2008] makes for a more original LXX text in Josh 24:30, 33.[51] The LXX has a longer text that adds information from Judges 2:6–10, 13; 3:7, 12, 14, and this information is used to reverse the order of Josh 24:29–31. Joshua is buried with flint knives imported from Joshua 5:2–3. After narrating the death of Joshua and Eleazar (24:33), the LXX 24:33b adds information from Judges 3:12, thus bypassing the two introductions in Judges 1:1-3:6, as well as Othniel, who in the MT is the ideal role model for a judge. [Tov, 2008: 49-50] translates the LXX addition back into Hebrew and adds references to parallels in Hebrew elsewhere:[52]

> And it happened after these things that Eleazar son of Aaron, the high priest, died, and was buried in Gabaath of Phinees his son, which he gave him in Mount Ephraim. + **On that day the sons of Israel took the ark of God and carried it around in their midst.** (cf v. 33 and Judg 20:28) **And Phinees served as priest in the place of Eleazar his father until he died, and he was interred in Gabaath, which was his town.** (Cf v. 28.) **And the sons of Israel departed each to their place and to their own city.** (Cf. Judg. 2:6, 12-13; 3:12-14.) **And the sons of Israel worshiped Astarte, and Ashtaroth, and the gods of nations around them. And the Lord delivered them into the hands of Eglon, the king of Moab and he dominated them eighteen years.**

The themes of the framework were discussed in [Winther-Nielsen, 1995: 293-294], but no display like figure 2 was provided for these verses. In a new textual corpus criticism, we would first exploit the text-level information contained in the ETCBC corpus. We would notice from the codes in the database that perfective *wayyiqtol* forms continues similar *wayyiqtol* forms in clauses (1)-(3). In Joshua 24:33 the code indicates that the conjunction is used before a *qatal* form which continues the preceding *wayyiqtol* and in turn is continued by *wayyiqtol*. From this it is clear that the last three clauses in the MT are usual run-of-the-mill closure information in Hebrew. In contrast, the LXX opens with a repetition of *wayhî ʔaḥᵃrê haddᵊvārîm hāʔēlleʰ* from the Hebrew of Josh 24:29, which is a superfluous repetition and without parallel in the discourse techniques for opening and closing segments in discourse in the Hebrew Bible. Furthermore, the MT of Josh 24:33a works well as a final wrapping up of the person gallery in the book, simply noting the death of the contemporary priest at from the time of Joshua. In contrast to this, the Astharoth gods in the Greek do not fit very well into the preceding account in Joshua after the covenant renewal, nor have they been mentioned before. It would also be difficult to explain the removal of the ideal judge Othniel with his family ties to Caleb. There is simply no need for such a surgical operation on the Hebrew text: the Greek ending can be explained as memory variants of a translator. He even recalls from memory the puzzling presence of the ark in Judges 20.

---

[49] Some scholars, e.g., [Carr, 2011: 74, n. 66] reject Auld's radical reconstruction that the late Book of Chronicles served as the basis for Joshua. Carr also rejects the assumption that Joshua 21 depends on 1 Chron 6:39-66 [ibid: 77-78 n. 71].

[50] The strongest critique of [Den Braber, 2010: 185-187] is that a rhetorical description of interclausal relation cannot be carried out successfully. Implicitly she overlooks the division of labour between the structural descriptions by the computer and the further work on functional grammar and interclausal grammar rhetorical structure theory.

[51] For discussion see [Nelson, 1997: 280–83], and especially [Butler, 2014: 335-336] who, in detailed discussion of current literature, argues that LXX harmonizes the MT.

[52] In the view of [Tov, 2008: 49], the beginning of Judges was missing in the *Urtext*, and such additions "point to the existence of a combined book of Joshua-Judges." One DSS witness lumps the ark, the deaths of Eleazar, Joshua, and the elders together with worship of Ashtaroth, thus indirectly confirming the Greek translation.

| | | | | |
|---|---|---|---|---|
| 1. | Way0 | 477 | 0•••••••• | ²⁹wayhî ʔaḥᵃrê haddᵊvārîm hāʔēlleʰ |
| | | | | and-it.was after the-events the-these |
| 2. | WayX | 477 | 1•••••••• | wayyāmot yᵊhôšuₐʕ bin-nûn ʕeved YHWH ben-mēʔāʰ wāʕeśer šānîm |
| | | | | and-he.died Joshua son.of-Nun servant.of YHWH son.of-100 and-10 years |
| 3. | Way0 | 477 | 2•••••••• | ³⁰wayyiqbᵊrû ʔōtô bigvûl naḥᵃlātô bᵊtimnat-seraḥ |
| | | | | and-they.buried him on-border his-lot in-Timnat-Serah |
| 4. | NmCl | 10 | 6••• | ʔᵃšer bᵊhar-ʔefrāyim miṣṣᵊfôn lᵊhar-gāʕaš |
| | | | | Which in-Mount-Ephraim from-south to-Mount-Ga'as |
| | | | | …. |
| 5. | XQt | 427 | 3•••••• | ³³wᵊʔelʕāzār ben-ʔahᵃrōn mēt |
| | | | | And-Eleazar son.of-Aaron he.died |
| 6. | Way0 | 472 | 4••••• | wayyiqbᵊrû ʔōtô bᵊgivʕat pînḥās bᵊnô |
| | | | | and-they.buried him in-Giv'at.of Pinhas his-son |
| 7. | xQt0 | 12 | 5•••• | ʔᵃšer nittan-lô bᵊhar ʔefrāyim |
| | | | | which he.gave-to.him in-Mount.of Ephraim |

Figure 2. Joshua 24:29-30.33 in display from Bible OL from [http19] [53]

From this it appears that there are no sufficient reasons for inventing a new ending for the book of Joshua on the basis of the Greek translation. In the end, however, we will have to consider each case of deviation individually, while we wait for better tools like Tiberias or for an analysis of translation shifts. When put to the test, the argument that we need to do good discourse-pragmatic work both on the Hebrew manuscripts and on the Greek translation still holds, especially when we allow for the evidence from memory variants.

For a new textual corpus criticism, the corpus-driven work enabled by the ETCBC database remains the most important model. Corpus technology will be much more helpful to students than an artificial *Urtext* or a redaction historical replacement. I would personally not want to give up the text of MT on such a fragile basis. However, corpus criticism and everything else boils down to a question of ultimate beliefs and the validity of the assumptions in our methods.

## Conclusions

The intertextuality of an ancient text has been explored from three different perspectives. The point of departure for a new kind of persuasive intertextuality is the corpus created by the ETCBC and a presentation of the Bible OL project that since 2008 has developed a corpus-driven learning environment. In this case, intertextuality means directing one's personal learning project through a persuasive interface of a linguistically annotated ancient text and using the force and flow of autonomy and mastery for accessing a socially relevant professional context. The review of the work with Bible OL gave me the opportunity to argue for extending this framework to a similar area of computer-assisted study of ancient texts. Selecting textual criticism as a related topic, I first looked at the advantages of using the commercially available Logos tools for teaching textual criticism. This kind of intertextuality typically uses a superb, but more traditional information technology. My last move was to look at ways of envisioning a new kind of corpus-based textual criticism. I suggest that there is great potential in exploring memory variants in a new application of the ETCBC database, and that we need a tool for analysis of translation shifts for the LXX. In the end I looked at Joshua to explore how the corpus works in a discussion of pros and cons of a very different ending in the Greek translation of Joshua. This last kind of intertextuality is the closest we can come to the use of our advanced tools for the highly scholarly tasks.

---

[53] A code like 477 indicates that a perfective *wayyiqtol* form (conjunction (4**) and *yiqtol* (*7*)) continues another *wayyiqtol* forms (**7). The code 427 indicates that the conjunction (4**) is used before a *qatal* (*2*) and continues *wayyiqtol* (**7). Other sigla like Way0 indicates a *wayyiqtol* without explicit subject, while WayX indicates the form has a subject. WXQt indicates a *qatal* verb form with conjunction and subject. 10 is a relative clause, of the NmCl variety (nominal clause or verbless clause), while 12 is a relative clause with *qatal*.

In the end I conclude that new open and persuasive technology has potential to gradually find its way around the globe, supporting a new kind of textual corpus criticism of the Hebrew Bible as well as all other tasks on cultural history, interpretation, theology, and education in the churches. Advanced modern commercial software like Logos offers first-rate information architecture integrated with many resources for a learner-friendly textual criticism. However, this software system, like any other system or print, can only increase the gap between the rich students in the West and students in the majority of the world who cannot afford textual criticism resources, either printed or digital. Without open-minded donors or new publishing solutions, books in libraries will continue to be the only solution for a while. The future lies in developing applications for corpora like the ETCBC and opening them to global access, so that the shortage of resources does not limit theological education. Finally, we can only hope that similar tools will be developed for Greek and Latin and for theological works and archives past and present.

# References 1: Editions and Digital Resources

Accordance: http://www.accordancebible.com/search/search.php#{"terms":"lxxg","page":1,"limit":20,"tab":"prod"}

Tov E. *The Parallel Aligned Hebrew-Aramaic and Greek Texts of Jewish Scripture.*
Logos edition 2003: https://www.logos.com/product/2209/the-parallel-aligned-hebrew-aramaic-and-greek-texts-of-jewish-scripture.

Pietersma A. and Wright B. *A new English translation of the Septuagint and other Greek translations traditionally included under that title.* Oxford University Press (Oxford), 2007.

**SP**

von Gall A. ed. *Der hebräische Pentateuch der Samaritaner*; Berlin 1966 (photomechanical reprint of Gießen 1914–1918), reprint. Walther de Gruyter (Berlin), 1993.

Tsedaka B. and Sullivan S. eds. *The Israelite Samaritan Version of the Torah: First English Translation Compared with the Masoretic Version*. Eerdmans (Grand Rapids), 2013.
Accordance: Samaritan Pentateuch with morphological tagging
http://www.accordancebible.com/store/details/?pid=SAMAR-T

**T**

McNamara M. ed. *The Aramaic Bible Series,* 22 volumes. Liturgical Press (Collegeville, Mn.), 1987-1997.
Logos edition: https://www.logos.com/product/31386/the-aramaic-bible-series
Accordance: Targums with morphological tagging
http://www.accordancebible.com/store/details/?pid=TARG-T

**S**

*Monographs of the Peshitta Institute Leiden*, 22 volumes. Brill (Leiden), 1972-.
Peshitta Institute Leiden.
Logos edition 2006: https://www.logos.com/product/4642/the-leiden-peshitta
Accordance: http://www.accordancebible.com/store/details/?pid=PESHOT-T

**V**

Weber R. and Gryson R., eds. *Biblia Sacra iuxta vulgatam versionem*. 5$^{th}$ edition. Deutsche Bibelgesellschaft (Stuttgart), 2007.
Logos version (editio minor, or 4$^{th}$ edition from 1994): https://www.logos.com/product/55086/stuttgart-scholarly-editions-old-and-new-testament#014
Accordance: http://www.accordancebible.com/store/details/?pid=PESHOT-T

# Bibliography 2: References